\pdfoutput=1

\documentclass[11pt]{article}
\usepackage[a4paper]{geometry}
\usepackage{clic2017}
\usepackage{times}
\usepackage{url}
\usepackage{latexsym}

\usepackage{soul}
\usepackage[dvipsnames]{xcolor}
\usepackage{array}
\newdimen\NetTableWidth

\title{NLP-CIC @ PRELEARN: Mastering prerequisites relations, from handcrafted features to embeddings\thanks{\ \ ``Copyright \textcopyright\
2020 for this paper by its authors. Use permitted under Creative Commons License Attribution 4.0 International (CC BY 4.0).''}}

  \author{ 
  Jason Angel \\
  Instituto Polit\'{e}cnico Nacional \\
  Mexico City, Mexico \\
  {\tt ajason08@gmail.com}
  \And
  Segun Taofeek Aroyehun \\
  Instituto Polit\'{e}cnico Nacional \\
  Mexico City, Mexico \\
  {\tt aroyehun.segun@gmail.com}
  \AND
  Alexander Gelbukh \\
  Instituto Polit\'{e}cnico Nacional \\
  Mexico City, Mexico \\
  {\tt www.gelbukh.com}
  }

\date{}

\begin{document}
\maketitle
\begin{abstract}
  \textbf{English.} 
 We present our systems and findings for the prerequisite relation learning task (PRELEARN) at EVALITA 2020. The task aims to classify whether a pair of concepts hold a prerequisite relation or not. We model the problem using handcrafted features and embedding representations for in-domain and cross-domain scenarios. 
Our submissions ranked first place in both scenarios with average F1 score of $0.887$ and $0.690$ respectively across domains on the test sets. We made our code freely available\footnote{\url{https://github.com/ajason08/EVALITA2020_PRELEARN}\label{code}}.
 \end{abstract}

\begin{abstract-alt}
 \textrm{\bf{Italiano.}} Conferiamo i nostri sistemi e risultati per il task di apprendimento dei relazioni di prerequisiti ``prerequisite relation learning" (PRELEARN), EVALITA 2020. Il task mira a classificare se una coppia di concetti c'è conessa tra una relazione di prerequisito. Modelliamo il problema utilizzando ingegneria delle caratteristiche e rappresentazione distribuita esaminando due modalità: in-domain e cross-domain.
 I nostri sistemi si sono classificati al primo posto con un F-score medio di $0,887$ e $0,690$ per le modalità di in-domain e cross-domain rispettivamente. 
 Inoltre, il nostro codice è open source code.

\end{abstract-alt}

\section{Introduction}

A prerequisite relation is a pedagogical relation that indicates the order in which concepts can be presented to learners. The relation can be used to guide the presentation sequence of topics and subjects during the design of academic programs, lectures, and curricula or instructional materials.

In this work, we present our systems to automatically detect prerequisite relations for Italian language in the context of the PRELEARN shared task \cite{PRELEARN2020} at EVALITA  2020 \cite{Evalita2020}. 
The evaluation of submissions considers: (1) in-domain and cross-domain scenarios defined by either the inclusion (in-domain) or exclusion (cross-domain) of the target domain in the training set. The four domains are 'data mining' (DM), 'geometry' (Geo), 'precalculus' (Prec), and 'physics' (Phy). (2) the type of resources (features) used to train the model -- raw text VS. structured information.


The combination of these settings defined the four PRELEARN subtasks. 
Formally, a prerequisite relation exists between two concepts if one has to be known beforehand in order to understand the other. 
For the PRELEARN task, given a pair of concepts, the relation exists only if the latter concept is a prerequisite for the former. 
Therefore, the task is a binary classification task.        

We approach the problem from two perspectives: handcrafted features based on lexical complexity and pre-trained embeddings. We employed static embeddings from Wikipedia and Wikidata, and contextual embeddings from Italian-BERT model.

\section{Related works}
Prerequisite relation learning has been mostly studied for the English language \cite{liang2018active,talukdar2012crowdsourced}. \newcite{adorni2019towards} performed unsupervised prerequisite relations extraction from textbooks using word co-occurrence and order of words appearance in the text. In the case of Italian language there is \textit{ITA-PREREQ} \cite{miaschi2019linguistically}, the first dataset for prerequisite learning, and actually the one used for the present work. It was automatically built as a projection of \textit{AL-CPL} \cite{liang2018active} from the English Wikipedia to the Italian Wikipedia. In addition, \newcite{miaschi2019linguistically} examines the utility of lexical features for individual concepts and features derived from the concept pairs.


\section{Methodology}
This section describes the data analysis, 
the features we used to model the task, and the system we finally submitted to the PRELEARN competition.

\subsection{Dataset}
The dataset provided by the organizers includes the concept pairs splitted into the following domains: 'data mining', 'geometry', 'precalculus' and 'physics'. 
The dataset contains the list of concepts with a link to the corresponding Wikipedia article. The first paragraph of such article is named the concept description.
All concept descriptions are cleaned in order to facilitate the extraction of information from the text, e.g. the mathematical expressions are already tagged 
using this pattern formula\_\textless number\textgreater.

Table \ref{dataset_table} displays the number of samples and the distribution over the prerequisite relations (positive samples) across domains for the training set. The test sets in turn exhibits a 50-50 distribution over positive and negative samples.

The only preprocessing we did was lowercase the concept description and remove line-breaks.

\begin{table}[h]
\begin{center}
\begin{tabular}{|l|c|c|c|c|}
\hline \bf  Domain & \bf Samples & \bf Prerequisites rel.\\ 
\hline
Data mining & 424&	0.257 \\
Geometry & 1548&	0.214\\
Precalculus & 2220&	0.142\\
Physics & 1716&	0.238\\
\hline
\end{tabular}
\end{center}
\caption{\label{dataset_table} Training set number of samples and distribution of prerequisite relations (positive samples) across domain}
\end{table}
\begin{table*}[ht]
\begin{center}
\begin{tabular}{|l|l|l||c|c|c|c|c|}
\hline \bf Scenario & \bf Resources & \bf System & \bf DM & \bf Geo & \bf Phy & \bf Prec & \bf AVG \\
\hline
in-domain & raw-text & Italian-BERT & 0.838 & 0.925 & 0.855 & 0.930 & 0.887\\
\hline
in-domain & structured & Complex+wd & 0.808 & 0.905 & 0.795 & 0.915 & 0.856\\
in-domain & structured & Complex & 0.828 & 0.895 & 0.785 & 0.885 & 0.848\\
\hline
cross-domain & raw-text & Italian-BERT & 0.565 & 0.785 & 0.635 & 0.775 & 0.690\\
\hline
cross-domain & structured & Complex+wd & 0.535 & 0.775 & 0.600 & 0.760 & 0.668\\
cross-domain & structured & Complex & 0.494 & 0.735 & 0.595 & 0.730 & 0.639\\
\hline
\end{tabular}
\end{center}
\caption{\label{results_test_table} Test set results for the four PRELEARN subtasks using F1-score}
\end{table*}

\begin{table*}[ht]
\begin{center}
\begin{tabular}{|c|c|c|}
\hline \bf Settings & \bf In-domain & \bf Cross-domain \\
\hline
raw-text & +2.1\% & +4.2\%  \\
structured & \textbf{+15.6\%} & \textbf{+4.8}\%  \\
\hline
\end{tabular}
\end{center}
\caption{\label{gap_result_table} Performance advantage over the next best participant on average across domains}
\end{table*}

\subsection{Features}
The following are the set of features we experiment with:
\par\textbf{Complexity-based:} a set of handcrafted features intended to measure how complex a concept is. The rationale is that less complex concepts are prerequisites for the more complex ones. We used some features that have been found effective for the task of complex word identification \cite{aroyehun2018complex}, specifically they are:
\begin{itemize}
    \item Age of acquisition of concept: we use \textit{ItAoA} \cite{montefinese2019italian}, a dataset of age of acquisition norms (we average the values for the different entries per word), to derive the age of acquisition for each concept we compute the geometric mean of values from ItAoA for words which occur in the concept description after replacing outliers (by the closest permitted value). In addition, we use the number of matches as a feature. 
    \item Age of acquisition of related concepts: 
    We derived a list of concepts related to each concept by matching which of them appears in the concept description. Then, we average the age of acquisition of those concepts. We also took the count of the related concepts.
    \item Description length: we count the number of words in the concept description.
    \item Number of mathematical expressions: we count the occurrence of  mathematical expressions. We assume that more complex concepts will have a higher occurrence of mathematical expressions in their descriptions. 
    \item Concept view frequency:
    the average of the daily unique visits by Wikipedia users (including editors, anonymous editors, and readers) over the last year. 
    We think that the number of visitors will be correlated with the degree of complexity of a concept. To gather this information we used the Pageviews Analysis of Wikipedia \footnote{\url{https://pageviews.toolforge.org}}. 
\end{itemize}

\par\textbf{Concept-to-Concept features:} they aim to model the relation between the concept pairs, specifically we evaluate whether a concept appears as a sub-string in the title or description of the other concept. We did this in both directions resulting in two features. We also represent the domain they belong to as a one-hot vector.

\par\textbf{Wiki-embeddings:} We map each concept identifier to their corresponding Wikipedia title and Wikidata identifier using the Wikidata Query Service\footnote{\url{query.wikidata.org}}. Then, we obtain the 100 dimensional vector for each Wikipedia title from a pre-trained Wikipedia embedding\footnote{\url{http://wikipedia2vec.s3.amazonaws.com/models/it/2018-04-20/itwiki_20180420_100d.pkl.bz2}} \cite{yamada2020wikipedia2vec}. Similarly, we use the Wikidata embedding\footnote{\url{https://dl.fbaipublicfiles.com/torchbiggraph/wikidata_translation_v1.tsv.gz}} \cite{pbg} to represent the Wikidata identifiers as 200 dimensional vectors.

\par\textbf{Italian-BERT features:} We used a pre-trained uncased version of Italian BERT (base model)\footnote{\url{https://huggingface.co/dbmdz/bert-base-italian-uncased}} provided by the MDZ Digital Library team (dbmdz) trained on 13GB of text mainly from Wikipedia and  other text sources. With this model, we get the 768 dimensional vector representation for a sequence corresponding to the [CLS] token as in the original implementation of BERT \cite{devlin-etal-2019-bert}. The sequence consists of the combination of the concept and its Wikipedia description.
\begin{table*}[ht]
\begin{center}

\begin{tabular}{|l|l|p{3cm}||c|c|c|c|c|}
\hline \bf Scenario & \bf Resources & \bf Feature set & \bf DM & \bf Geo & \bf Phy & \bf Prec & \bf AVG \\
\hline
in-domain & raw & complexity & 0.646 & 0.817 & 0.622 & 0.792 & 0.720\\
in-domain & raw & wp\_embedding & 0.705 & \textbf{0.818} & 0.670 & 0.827 & 0.755\\
in-domain & raw & Italian-BERT & \textbf{0.947} & 0.746 & \textbf{0.829} & \textbf{0.842} & \textbf{0.841}\\
\hline
in-domain & structured & complexity +page\_view & 0.648 & 0.805 & 0.629 & 0.804 & 0.721\\
in-domain & structured & wd\_embedding & 0.660 & 0.814 & 0.674 & 0.838 & 0.746\\
in-domain & structured & wd+wp\_embedding & 0.694 & \textbf{0.824} & 0.672 & 0.831 & 0.755\\
in-domain & structured & complexity +page\_view +wd\_embedding & \textbf{0.697} & 0.823 & \textbf{0.686} & \textbf{0.845} & \textbf{0.763}\\
\hline
cross-domain & raw & complexity & 0.072 & 0.592 & 0.258 & \textbf{0.586} & 0.377\\
cross-domain & raw & wp\_embedding & 0.000 & 0.622 & 0.079 & 0.344 & 0.261\\
cross-domain & raw & Italian-BERT & \textbf{0.145} & \textbf{0.646} & \textbf{0.460} & 0.570 & \textbf{0.455}\\
\hline
cross-domain & structured & complexity +page\_view & \textbf{0.107} & 0.588 & 0.297 & 0.577 & 0.392\\
cross-domain & structured & wd\_embedding & 0.000 & \textbf{0.661} & 0.355 & 0.608 & 0.406\\
cross-domain & structured & wd+wp\_embedding & 0.000 & 0.660 & 0.332 & 0.605 & 0.399\\
cross-domain & structured & complexity +page\_view +wd\_embedding & 0.064 & 0.645 & \textbf{0.366} & \textbf{0.630} & \textbf{0.426}\\
\hline
\end{tabular}
\end{center}
\caption{\label{results_dev_table} Ablation analysis results using F1-score (validation set for Italian-BERT and 10-fold for the others)}
\end{table*}

\subsection{Systems}

Considering the proposed features and our experimental results at Section \ref{ablation}, we proposed the following three systems to address both, in-domain and cross-domain scenarios. 
For the in-domain scenario we trained with a combination of all the training samples per domain. In the same way, we combined the remaining three domains for each cross-domain experiment (i.e. excluding samples from the target domain).

\par\textbf{Complex:} a completely handcrafted machine learning system, it uses all the complexity-based and Concept-to-Concept features (except the domain vector for cross-domain scenario), and we normalize the features using Z-score normalization. This system uses a tree-ensemble learner as classifier\footnote{Other classifiers were tested and obtained lower peformance} with the default parameters provided by \newcite{breiman2001random}\footnote{\url{https://cran.r-project.org/web/packages/randomForest/index.html}}. This system participated under the structured resource setting because the ``concept  view  frequency" feature is structured information.

\par\textbf{Complex+wd:} an improved version of the \textit{Complex} system by only concatenating the Wikidata embedding of each concept in the concept pair to the feature set. This system participated under the structured resource setting as well. We decided to not include the Wikipedia embeddings considering the ablation analysis we present in Table \ref{results_dev_table}.

\par\textbf{Italian-BERT:} a single layer neural network mapping the 768 features from the [CLS] to the output space of dimension $2$ as a sequence pair classification task. In addition, the pre-trained weights of the base model are fine-tuned on the training dataset. We fine-tune the base model using the huggingface transformers library (version 3.1) for Pytorch \cite{Wolf2019HuggingFacesTS}.
In the in-domain scenario, we use the following training parameters: the number epochs is $10$, learning rate is $5e-5$, weight decay is $0.01$, batch size is $32$, warm up steps is $100$, optimizer is AdamW with a linear schedule after a period of warm up steps.
We find that the model exhibits high variance across runs in our cross-domain experiments. Hence, in addition to the parameter settings for the in-domain experiments, we choose the number of training steps using a validation set for the unseen target domain. Accordingly, we set the maximum training step to $400$ and the warm up steps to $100$, $200$, $150$, and $200$ for data mining, geometry, physics, and precalculus cross-domain scenarios respectively.


\section{Results}
Table \ref{results_test_table} shows our per-domain results for our systems indicating the kind of scenario and resources they used. We observe the clear superiority of Italian-BERT which only relies on raw-text resources. This suggest that just fine-tuning BERT is enough for gaining a notion of prerequisite relations on concepts. Still, the systems based on handcrafted features and non-contextual embedding exhibit competitive results, with a good enough performance to rank first in the structured resource setting, while being faster, more interpretable and simpler than the Italian-BERT counterpart. 

The results showed that there is a huge performance reduction for the cross-domain scenario. The largest performance drop is on the ``data mining" domain. Given that we train our models on the combination of examples from all other domains, it is likely that the probable cause is the domain mismatch. Yet, the reduction on the test sets are smaller than what we observe in our K-fold experiments and validation sets.

In addition, we show in Table \ref{gap_result_table} the performance advantage we obtained over the next best participant based on the ranking released by the organizers. 

One can see that the greater performance advantage is from the structured resource setting. This suggests that the ``Concept view frequency" and the Wikidata embedding features are effective.

\section{Discussion: ablation analysis}\label{ablation}
During the creation our systems we perform several experiments over the possible features to use. We did 10-fold cross validation for the in-domain experiments except with the Italian-BERT\footnote{Due to its high computational requirements}, for which we used a stratified split of 30\% for validation set. Table \ref{results_dev_table} shows the experimental results over the training (validation) set for both, in-domain and cross-domain scenarios. The ``Resources" column serves to identify the type of resources used for the current feature. 

We observe that the ``data mining" domain appears to be difficult in the cross-domain scenario, models based on the non-contextual embedding features obtain results of zero. We suspect that this difficulty is due to the domain mismatch.


Based on these results, we select the Italian-BERT for the raw-text setting, and the ``complexity +page\_view" and the addition of Wikidata embeddings (``wd\_embedding") for the structured resource setting for our submissions.






\section{Conclusion}
We tackle the task of prerequisite relation learning using a variety of systems that explore three set of features: handcrafted features based on complexity intuitions, embedding models from Wikipedia and Wikidata, and contextual embedding from Italian-BERT model. We examine the capabilities of our models in in-domain and cross-domain scenarios. 
Our models ranked first in all the subtask of the PRELEARN competition at EVALITA 2020. We found that although our Italian-BERT model outperformed the others, the simpler models show competitive results.

We plan to further examine the impact of using a combination of all possible domains as training set on the performance of our models.
\section*{Acknowledgments}

The authors thank CONACYT for the computer resources provided through the INAOE Supercomputing Laboratory's Deep Learning Platform for Language Technologies.

\bibliographystyle{acl}
\bibliography{references}

\end{document}